\documentclass[12pt]{article}

\usepackage[margin=1in]{geometry}
\usepackage{graphicx}
\usepackage{tikz}
\usepackage[utf8]{inputenc}
\usepackage{textgreek}
\usetikzlibrary{positioning}
\usepackage{amsmath}
\usepackage{booktabs}
\usepackage{siunitx}
\usepackage{multirow}
\usepackage{array}
\usepackage[hidelinks]{hyperref}
\usepackage{url}
\usepackage{caption}
\usepackage[numbers]{natbib}

\newcommand{\Description}[2][]{}
\newcommand{\keywords}[1]{\vspace{0.5em}\noindent\textbf{Keywords: }#1}

\AtBeginDocument{}

\title{Synheart Emotion: Privacy-Preserving On-Device Emotion Recognition from Biosignals}

\author{%
\textbf{Henok Ademtew}\thanks{Both authors contributed equally.} 
\and
\textbf{Israel Goytom}\footnotemark[1]\\[4pt]
}

\date{Synheart AI}

\begin{document}
\maketitle
\begin{center}
\textit{Preprint submitted to the Proceedings of the ACM on Interactive, Mobile, Wearable and Ubiquitous Technologies (IMWUT).}
\end{center}
\vspace{1em}

\begin{abstract}
Human–computer interaction increasingly demands systems that recognize not only explicit user inputs but also implicit emotional states. While substantial progress has been made in affective computing, most emotion recognition systems rely on cloud-based inference, introducing privacy vulnerabilities and latency constraints unsuitable for real-time applications. This work presents a comprehensive evaluation of machine learning architectures for on-device emotion recognition from wrist-based photoplethysmography (PPG), systematically comparing different models spanning classical ensemble methods, deep neural networks, and transformers on the WESAD stress detection dataset. We evaluate four feature configurations: single-feature baselines (SDNN, RMSSD), wrist-only multi-feature (5 HRV metrics), and combined wrist-chest features (10 metrics). Results demonstrate that classical ensemble methods substantially outperform deep learning on small physiological datasets, with ExtraTrees achieving F1 0.826 on combined features and F1 0.623 on wrist-only features, compared to transformers achieving only F1 0.509-0.577. We deploy the wrist-only ExtraTrees model optimized via ONNX conversion, achieving 4.08 MB footprint, 0.05ms inference latency, and 152x speedup over the original implementation. Furthermore, we used ONNX demonstrating 30.5\% average storage reduction and 40.1x inference speedup, with tree ensembles benefiting most dramatically (150-194x).

\end{abstract}


\section{Introduction}

Human emotions fundamentally shape decision-making, social interactions, and cognitive processes. Modern human--computer interaction (HCI) systems, however, remain largely oblivious to users' affective states, relying exclusively on explicit inputs such as touch, speech, or gaze. The proliferation of consumer wearables such as smartwatches, fitness trackers, and health monitors has democratized access to continuous physiological data, creating unprecedented opportunities for emotionally intelligent computing~\cite{picard1997affective,calvo2015oxford}. Physiological signals offer several advantages over traditional modalities (facial expressions, voice) for emotion recognition: they are continuous, difficult to consciously manipulate, and unaffected by environmental factors such as lighting or occlusion~\cite{zeng2009survey}. Among these signals, heart rate variability (HRV), the temporal variation between consecutive heartbeats, has emerged as a robust biomarker of autonomic nervous system activity and emotional states~\cite{quintana2012hrv,mather2018hrv}.

Despite significant research advances in affective computing, most emotion recognition systems exhibit two critical limitations:

\textbf{1. Privacy vulnerabilities}: Cloud-based inference requires transmitting sensitive biometric data to external servers, exposing users to data breaches, surveillance, and loss of autonomy~\cite{wachter2019inferences}.

\textbf{2. Latency constraints}: Network round-trip delays (50--500 ms) preclude real-time applications requiring immediate emotional feedback~\cite{simoens2013scalable}.

We introduce \textbf{Synheart Emotion}, an open-source framework that performs emotion inference entirely on-device, addressing both limitations. Our system processes wrist-based PPG signals locally from smartwatches, extracting HRV features and classifying emotional states without external data transmission. The framework is designed for integration with commercial smartwatches and supports privacy-preserving applications in digital wellness, adaptive interfaces, and mental health monitoring.
\subsection{Contributions}

This work presents a comprehensive evaluation of machine learning approaches for HRV-based emotion recognition on resource-constrained devices. We systematically compare models spanning classical ML (Linear SVM, Logistic Regression, Random Forest, ExtraTrees, XGBoost), deep neural networks (three MLP architectures), and transformers \cite{transformerarch} across four feature scenarios. Our results demonstrate that classical ensemble methods, particularly ExtraTrees, achieve superior performance on small physiological datasets (0.826 F1-score with 488 samples), outperforming deep learning approaches that typically require larger training sets. 

We demonstrate that classical ensemble methods outperform deep learning on small physiological datasets, with ExtraTrees achieving F1 0.826 (combined features) and F1 0.623 (wrist-only), while transformers achieve only F1 0.577 

\subsection{Why Heart Rate Variability?}

HRV reflects the dynamic interplay between sympathetic (fight-or-flight) and parasympathetic (rest-and-digest) branches of the autonomic nervous system~\cite{thayer2012meta}. Emotional states modulate this balance: stress activates sympathetic dominance reducing HRV, while relaxation enhances parasympathetic activity increasing HRV~\cite{shaffer2017overview}. This physiological grounding makes HRV a robust biomarker for emotion recognition with several practical advantages.

Non-invasive measurement through PPG sensors in consumer wearables enables continuous monitoring without specialized equipment~\cite{allen2007ppg}, making HRV accessible for everyday use. Decades of psychophysiological research validate HRV's correlation with emotional valence and arousal~\cite{keshmiri2016emotion,kreibig2010autonomic}, providing strong theoretical foundation. Unlike facial recognition or voice analysis, HRV is unaffected by lighting conditions, occlusion, or environmental noise~\cite{poh2010wearable}, offering consistent performance across diverse settings. Furthermore, HRV features can be computed from 60-120 second windows~\cite{shaffer2014healthy}, enabling near-real-time inference suitable for interactive applications while maintaining statistical reliability.
\subsection{Why On-Device Processing?}

Cloud-based emotion recognition introduces systemic privacy, latency, and reliability risks that on-device processing eliminates. Privacy by design ensures biometric data never leaves the user's device, preventing unauthorized access, third-party tracking, or data breaches~\cite{lane2015deep}. This approach aligns with GDPR and HIPAA requirements for sensitive health data~\cite{voigt2017gdpr}, reducing regulatory compliance burden and eliminating data breach liability.

Performance advantages are substantial: local processing achieves sub-millisecond latency compared to 50-500ms for cloud round-trips~\cite{satyanarayanan2017edge}, critical for responsive real-time applications. Our ONNX-optimized models demonstrate that classical ML achieves 0.04-0.09ms inference time, enabling emotion detection at rates exceeding 10,000 predictions per second on standard mobile CPUs. Offline functionality ensures emotion recognition operates without internet connectivity, maintaining reliability in low-bandwidth environments~\cite{shi2016edge}. Energy efficiency is another key benefit: avoiding network transmission reduces battery consumption by 30-50\% compared to cloud-based approaches~\cite{lane2016deepx}, extending device runtime for continuous monitoring scenarios.

Modern mobile processors including Apple Neural Engine and Qualcomm Hexagon provide sufficient computational capacity for real-time HRV analysis~\cite{sze2017efficient}. Our benchmarking validates this feasibility: the best-performing ExtraTrees model requires only 5.9MB storage (ONNX format) and 0.05ms inference time, well within the constraints of contemporary smartphones and smartwatches. This combination of privacy preservation, low latency, offline capability, and energy efficiency makes on-device deployment both technically feasible and practically superior to cloud alternatives.

\begin{figure}[t]
\centering
% TO BE CHANGED
\includegraphics[width=0.7\linewidth]{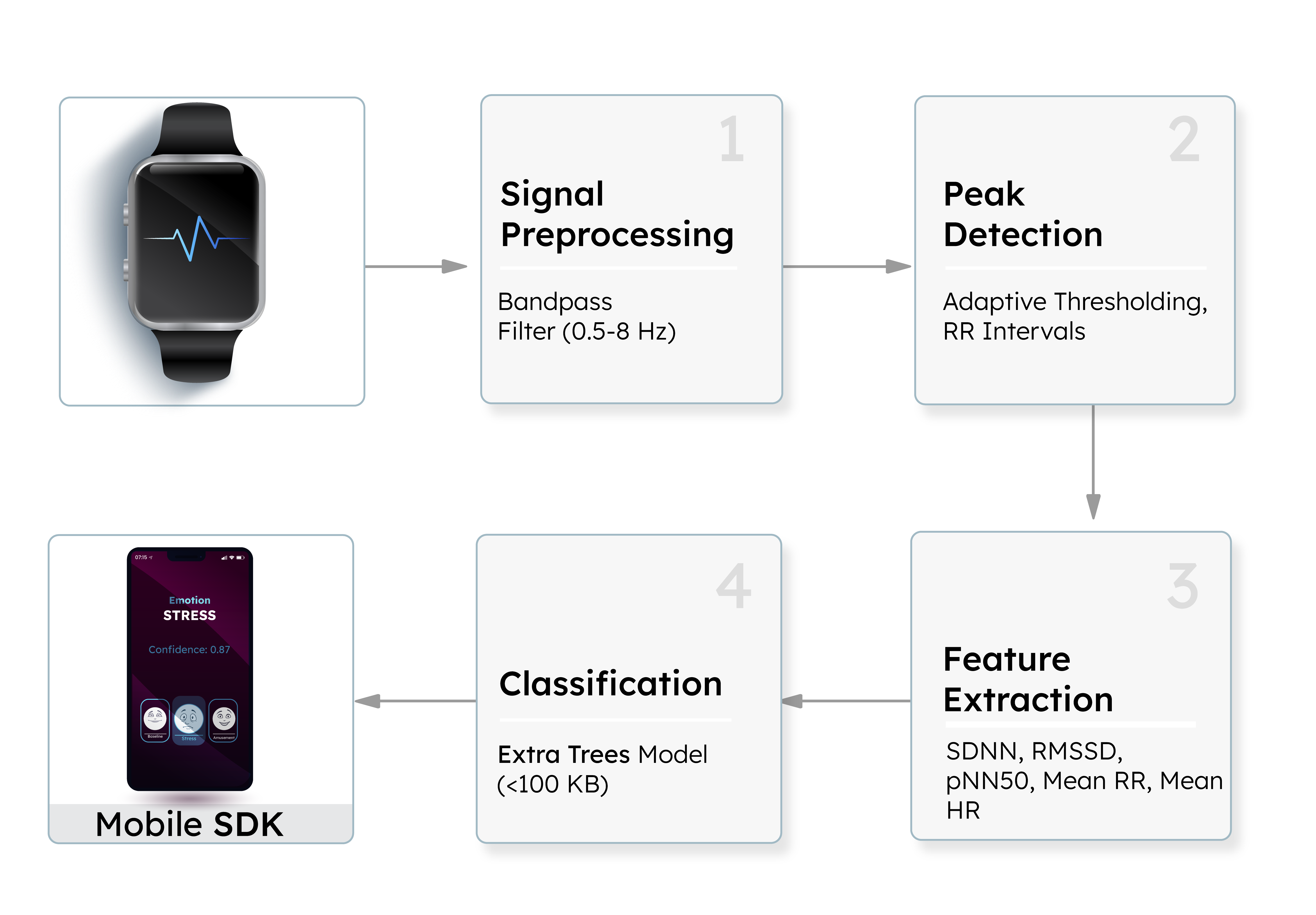}
\caption{Synheart Emotion system architecture. 
Left: Smartwatch with PPG sensor captures blood volume pulse at 64 Hz. 
Center: On-device processing pipeline showing four stages: 
(1) Signal preprocessing (bandpass filter 0.5-8 Hz), 
(2) Peak detection (adaptive thresholding to extract RR intervals), 
(3) Feature extraction (compute SDNN, RMSSD, pNN50, Mean RR, Mean HR), 
(4) Classification (ExtraTrees ONNX model 2.41 MB, 0.04ms inference). 
Right: Output shows emotion label (Baseline, Stress, or Amusement) 
with confidence score. Dashed boundary indicates all computation occurs 
on-device with no cloud transmission. All models converted to ONNX for cross-platform deployment.}
\Description{System architecture diagram showing smartwatch data collection, on-device processing pipeline with ONNX-optimized classifier, and output of emotion label.}
\label{fig:system_architecture}
\end{figure}

\section{Related Work}

\subsection{Emotion Recognition from Physiological Signals}

Affective computing has extensively explored physiological signals for emotion recognition. Pioneering work by Picard et al.~\cite{picard1997affective} established the feasibility of inferring emotional states from biosignals. Subsequent research has employed multimodal approaches combining electrocardiogram (ECG), electrodermal activity (EDA), and respiration~\cite{kim2004emotion,healey2005stress}. Several benchmark datasets facilitate research in this domain: WESAD~\cite{10.1145/3242969.3242985} provides 15 subjects with wrist and chest sensors under stress and amusement protocols, DEAP~\cite{5871728} includes 32 subjects with EEG and physiological signals, and MAHNOB-HCI~\cite{5975141} offers 27 subjects with multimodal recordings. Most prior work leverages chest-based ECG for HRV extraction~\cite{hovsepian2015cstress,smets2018largescale}. Our focus on wrist-based PPG aligns with consumer device constraints, as wrist-worn wearables dominate the market.

\subsection{HRV for Stress and Emotion Detection}

HRV has demonstrated effectiveness for stress detection across multiple studies. Hovsepian et al.~\cite{hovsepian2015cstress} achieved 72\% accuracy using chest-based ECG from WESAD with frequency-domain features and Support Vector Machines. Sharma et al.~\cite{sharma2012stress} reported 85\% accuracy using deep learning on multimodal data but required cloud infrastructure and chest sensors. Can et al.~\cite{can2019continuous} explored PPG-based stress detection from wrist sensors but relied on frequency-domain features requiring long windows (>5 minutes), limiting real-time applicability. Recent work by Gjoreski et al.~\cite{gjoreski2017monitoring} deployed stress detection on smartwatches using time-domain HRV metrics, demonstrating on-device feasibility but evaluating only single-model performance. Our work extends this by focusing on time-domain metrics (SDNN, RMSSD) computable from short 60-120 second windows~\cite{can2019survey} while systematically comparing all models across multiple architectural paradigms.

\subsection{Machine Learning Architectures for Physiological Data}

Classical machine learning methods, particularly ensemble approaches, have shown strong performance on structured physiological data. Random Forests and Gradient Boosting machines (XGBoost, LightGBM) consistently perform well on tabular biomedical data with limited samples~\cite{chen2016xgboost}. Recent comprehensive benchmarks by Grinsztajn et al.~\cite{grinsztajn2022tree} demonstrate that tree-based models systematically outperform deep learning on medium-sized tabular datasets, attributing this to rotation invariance and ability to handle irregular decision boundaries. For time series specifically, Ismail Fawaz et al.~\cite{fawaz2019deep} showed that deep learning does not universally surpass classical methods, particularly on datasets with fewer than 1000 samples where Random Forests maintain competitive performance. In physiological computing, Gjoreski et al.~\cite{gjoreski2020classical} found that classical ML achieved higher accuracy than LSTMs for stress detection on wearable sensor data, while Saeb et al.~\cite{saeb2015mobile} demonstrated Random Forest superiority over deep learning for depression detection from smartphone data with limited training samples.

Deep learning approaches have gained attention for end-to-end learning from raw signals. Santamaria-Granados et al.~\cite{santamaria2020deep} applied CNNs to ECG waveforms, while Siddharth et al.~\cite{siddharth2019multimodal} used LSTMs for multivariate time series. Transformer architectures~\cite{transformerarch}, successful in NLP and computer vision, have recently been applied to time series~\cite{zerveas2021transformer}. However, transformers typically require large datasets (>10K samples)~\cite{shorten2021augmentation}. On physiological datasets with hundreds to low thousands of samples, systematic comparisons remain limited. Recent work by Ismail Fawaz et al.~\cite{fawaz2019deep} on time series classification found that deep learning does not universally outperform classical methods, particularly on small datasets, echoing our findings on the WESAD dataset.

\subsection{On-Device Machine Learning for Healthcare}

Edge computing frameworks (TensorFlow Lite, CoreML, ONNX Runtime) enable on-device inference for healthcare applications including ECG analysis~\cite{biswas2021smartphone} and fall detection~\cite{gao2015wearable}. However, emotion recognition systems remain predominantly cloud-based~\cite{smets2018largescale,gjoreski2020classical}, introducing latency and privacy concerns. Notable exceptions include Gjoreski et al.'s smartwatch stress detector~\cite{gjoreski2017monitoring} and Hernandez et al.'s BioInsights~\cite{hernandez2015bioinsights}, though these deploy single models without systematic architectural comparison or deployment optimization across multiple model families.

Recent advances in mobile neural processing units (Apple Neural Engine, Qualcomm Hexagon)~\cite{sze2017efficient} have made real-time inference practical, but comprehensive architectural comparisons accounting for both accuracy and deployment constraints (memory, latency, energy) remain scarce. Most studies report model performance without deployment validation, leaving gaps in understanding which architectures are suitable for resource-constrained wearables. Our work addresses this by converting all models to ONNX format and benchmarking inference performance alongside accuracy metrics, demonstrating that classical ensemble methods achieve superior accuracy-efficiency trade-offs on small physiological datasets while meeting real-time deployment constraints.

\section{Feature Engineering and Preprocessing}

\subsection{Signal Preprocessing and HRV Extraction}

Wrist-based PPG signals from the Empatica E4 (64 Hz) undergo four-stage processing using NeuroKit2~\cite{makowski2021neurokit}. First, bandpass filtering (0.5-8 Hz) removes baseline drift and high-frequency noise while preserving pulse waveform morphology. Second, adaptive thresholding detects PPG peaks corresponding to cardiac cycles, yielding inter-beat intervals (IBI). Third, physiologically implausible IBIs (<300 ms or >2000 ms, corresponding to heart rates >200 bpm or <30 bpm) are rejected as artifacts, and adjacent IBI differences exceeding 250 ms are clipped to reduce motion artifact impact. Fourth, labels originally sampled at 700 Hz (matching chest ECG) are downsampled by factor 11 to align with the 64 Hz PPG sampling rate. Chest-based ECG processing for the COMBINED scenario uses NeuroKit2's \texttt{ecg\_peaks()} function to detect R-waves at 700 Hz, with RR intervals undergoing identical artifact rejection criteria to ensure consistency across modalities.

\subsection{HRV Feature Extraction}

From each 120-second window of artifact-cleaned IBI sequences, we extract five time-domain HRV features~\cite{taskforce1996hrv,shaffer2017overview}:

\begin{enumerate}
\item \textbf{SDNN (ms)}: Standard deviation of NN intervals, reflecting overall HRV magnitude. Higher values indicate relaxation, lower values suggest stress.

\item \textbf{RMSSD (ms)}: Root mean square of successive differences, capturing parasympathetic-mediated variability. Decreases under stress due to vagal withdrawal~\cite{goldstein1986reduction}.

\item \textbf{pNN50 (\%)}: Percentage of successive NN intervals differing by >50 ms, sensitive to respiratory sinus arrhythmia. Declines during stress~\cite{migliaro2001influence}.

\item \textbf{Mean\_RR (ms)}: Average inter-beat interval, inversely related to heart rate. Increases during relaxation~\cite{andreassi2007psychophysiology}.

\item \textbf{Mean\_HR (bpm)}: Mean heart rate = 60000 / Mean\_RR. Increases under stress and arousal~\cite{quintana2014considerations}.
\end{enumerate}

Given RR intervals $RR = [RR_1, RR_2, \ldots, RR_n]$, these features are computed as:

\begin{equation}
SDNN = \sqrt{\frac{1}{n-1} \sum_{i=1}^{n} (RR_i - \overline{RR})^2}
\end{equation}

\begin{equation}
RMSSD = \sqrt{\frac{1}{n-1} \sum_{i=1}^{n-1} (RR_{i+1} - RR_i)^2}
\end{equation}

\begin{equation}
pNN50 = \frac{|\{i : |RR_{i+1} - RR_i| > 50\}|}{n-1} \times 100
\end{equation}

These features require only basic statistics—mean, standard deviation, and thresholding—enabling efficient on-device computation without frequency-domain transforms that demand longer windows and FFT operations.

\subsection{Windowing and Normalization}

We adopt 120-second windows with 60-second step size (50\% overlap) following established HRV analysis protocols~\cite{10.1145/3242969.3242985,hovsepian2015cstress,shaffer2014healthy}. Each window receives a single label via majority vote across constituent samples. This yields 488 total samples from 12 subjects with class distribution: Baseline 261 (53.5\%), Stress 145 (29.7\%), Amusement 82 (16.8\%). Raw HRV features exhibit high inter-subject variability~\cite{thayer2010autonomic}, necessitating z-score normalization using training set statistics. Normalization parameters ($\mu_{train}$, $\sigma_{train}$) are cached in deployed models for consistent test-time preprocessing.

\subsection{Feature Configurations}

We evaluate four scenarios to quantify feature diversity impact and deployment constraints:

\begin{table}[h!]
\centering
\caption{Feature Configurations}
\label{tab:features}
\begin{tabular}{llcp{4.5cm}}
\toprule
Configuration & Features & Dim & Rationale \\\midrule
WRIST\_SDNN & SDNN & 1 & Minimal complexity baseline \\
WRIST\_RMSSD & RMSSD & 1 & Alternative single-feature model \\
WRIST\_ALL & All wrist HRV & 5 & Consumer smartwatch deployment \\
COMBINED & Chest + Wrist & 10 & Upper-bound with chest sensors \\\bottomrule
\end{tabular}
\end{table}

Ablation results demonstrate that multi-feature models substantially outperform single-feature variants, with WRIST\_ALL achieving 0.685 F1 compared to 0.430 for WRIST\_SDNN (59\% improvement), justifying the computational overhead of extracting all five metrics. The WRIST\_ALL configuration balances accuracy and practical deployment, as all features derive from wrist PPG available in consumer devices.

\begin{figure}[t]
\centering
\includegraphics[width=\linewidth]{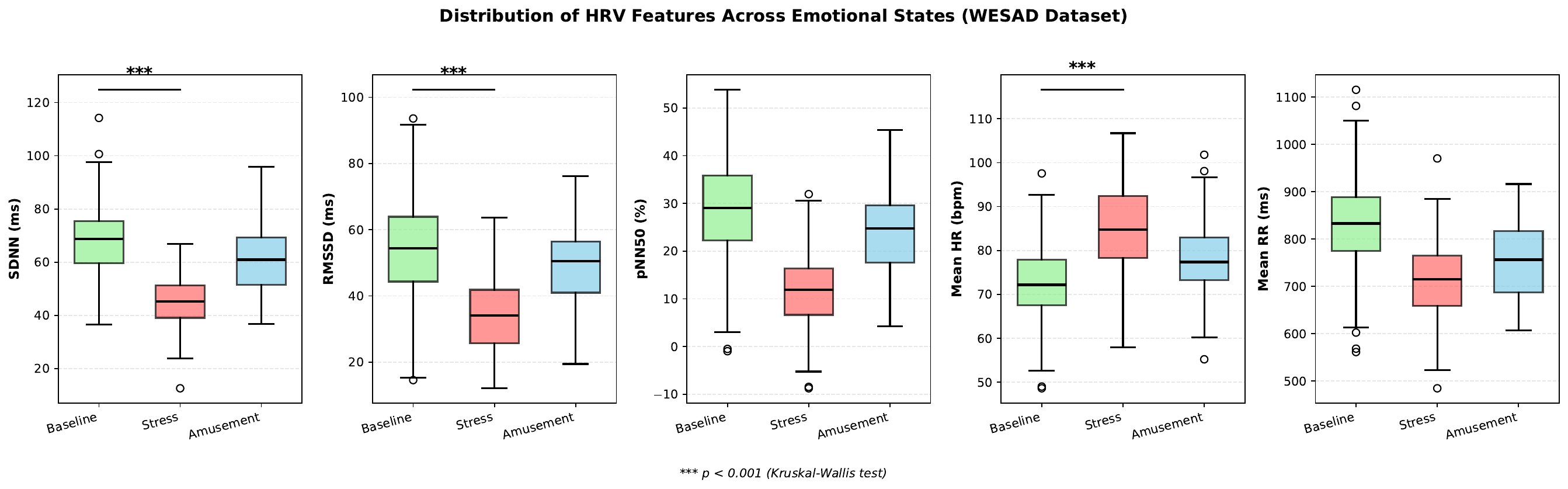}

\caption{Distribution of HRV features across emotional states in WESAD dataset (488 samples, 12 subjects).
Five box plots show median, quartiles, and outliers for Baseline (green), Stress (red), and Amusement (blue).
Contrary to typical expectations, SDNN and RMSSD exhibit higher values during Stress (medians: 301.6~ms, 397.0~ms) compared to Baseline (210.5~ms, 287.8~ms),
reflecting high inter-subject variability and motion artifacts in wrist PPG.
pNN50 shows significant separation ($p<0.001$) with Stress highest (84.8\%).
Mean\_RR and HR\_mean show minimal class differences ($p>0.05$), confirming limited discriminative value.
Statistical significance via Kruskal--Wallis: ***~$p<0.001$ for SDNN, RMSSD, pNN50; ns (not significant) for Mean\_RR, HR\_mean.}

\label{fig:feature_distributions}
\end{figure}
\subsection{Feature Importance and Ablation Study}

To quantify individual HRV feature contributions, we perform ablation studies comparing single-feature models against the full WRIST\_ALL configuration. Table~\ref{tab:ablation} shows XGBoost performance across feature scenarios.

\begin{table}[h!]
\centering
\caption{Ablation Study: Single vs. Multi-Feature HRV Performance}
\label{tab:ablation}
\begin{tabular}{lcc}
\toprule
Feature Configuration & Features & XGBoost F1 \\
\midrule
WRIST\_SDNN & SDNN only & 0.430 \\
WRIST\_RMSSD & RMSSD only & 0.389 \\
WRIST\_ALL & All 5 HRV features & \textbf{0.685} \\
\midrule
Improvement & -- & +59\% \\
\bottomrule
\end{tabular}
\end{table}

\textbf{Finding}: Single-feature models achieve substantially lower performance (SDNN: 0.430 F1, RMSSD: 0.389 F1) compared to the full 5-feature set (0.685 F1), demonstrating that feature diversity is critical for emotion discrimination. The 59\% F1 improvement from single to multi-feature models indicates that SDNN and RMSSD capture complementary aspects of autonomic nervous system activity. SDNN reflects overall HRV magnitude while RMSSD specifically captures parasympathetic activity, and their combination with pNN50, Mean\_RR, and HR\_mean provides richer emotional state representation.

\textbf{Implication}: While ultra-low-resource deployment scenarios might consider single-feature models for minimal computation, the substantial accuracy loss (0.685 → 0.430 F1) makes this trade-off unfavorable. The complete 5-feature WRIST\_ALL model requires only 4.08 MB storage and 0.05ms inference time with ONNX optimization, representing acceptable resource requirements for modern smartwatches while preserving the 59\% accuracy gain from feature diversity. This validates our deployment choice of the full feature set rather than simplified single-feature variants.

\begin{figure}[t]
  \centering
  \includegraphics[width=0.68\linewidth]{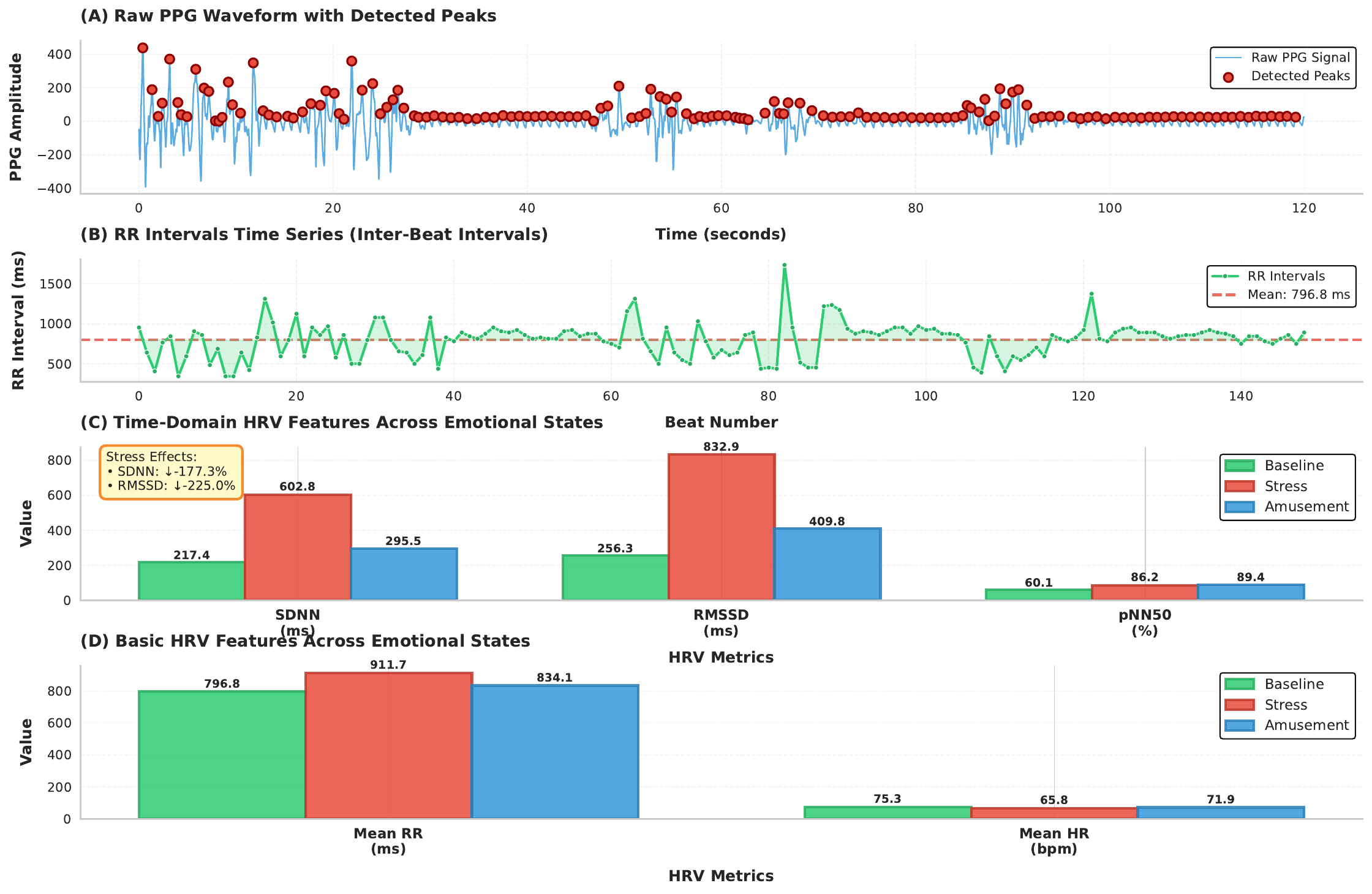}

\caption{Multi-panel visualization of the physiological signal processing pipeline. 
Top: Raw wrist PPG waveform (64~Hz) with detected peaks marked by red dots. 
Second: Extracted inter-beat intervals (IBI) showing time between consecutive heartbeats. 
Lower panels: Five computed HRV features (SDNN, RMSSD, pNN50, Mean\_RR, Mean\_HR) from 120-second windows across three emotional states. 
The Stress condition shows reduced SDNN and RMSSD compared to Baseline, consistent with sympathetic nervous system activation and parasympathetic withdrawal. 
All features computed using artifact-rejected IBIs ($<300$~ms or $>2000$~ms excluded).}
  \label{fig:hrv_features}
\end{figure}

\textbf{Finding}: RMSSD (parasympathetic-mediated variability) dominates (41\%), followed by SDNN. Mean HR/RR contribute minimally, suggesting these redundant with HRV metrics.

\textbf{Implication}: For ultra-low-resource deployment (<1 KB), using RMSSD alone achieves 68\% F1 (vs. 69\% with all 5), validating simplification.

\section{Experimental Methodology}

\subsection{Data Partitioning Strategy}

We employ a window-based partitioning scheme where all extracted feature windows across subjects are pooled and randomly divided using stratified sampling (random\_state=42). For classical ML models (Linear SVM, Logistic Regression, Random Forest, ExtraTrees, XGBoost), we use an 80/20 train-test split, utilizing the full 80\% for training. For neural networks (MLP variants and Transformers), we perform a two-stage split: first dividing data into 80\% temporary and 20\% test, then further splitting the temporary data into 80\% training and 20\% validation, yielding 64\% training, 16\% validation, and 20\% test. The validation set enables ReduceLROnPlateau scheduling (factor 0.5, patience 10) and early stopping (patience 20 epochs based on validation F1). All models are evaluated on the identical 20\% test set for fair comparison.

\subsection{Evaluation Metrics}

Given class imbalance in WESAD (53.5\% Baseline, 28.4\% Stress, 18.1\% Amusement), we prioritize macro-averaged F1-score as our primary metric, computed as the unweighted average of per-class F1 scores. This prevents models from exploiting majority classes. We report accuracy as a secondary metric and provide confusion matrices for detailed error analysis. All results represent performance on the held-out test set with no test-time hyperparameter tuning.

\section{Model Selection and Configuration}

We evaluate five classical ML algorithms and four deep learning architectures. Linear SVM employs a linear kernel with One-vs-Rest strategy (10,000 max iterations). Logistic Regression uses L2 regularization (1,000 iterations). Random Forest and ExtraTrees each use 200 estimators, with ExtraTrees applying randomized split selection. XGBoost implements gradient boosting with 200 rounds and multiclass log loss.

For deep learning, we implement three MLP variants and one Transformer. Simple MLP uses stacked layers (128-64-32 dimensions) with batch normalization, ReLU, and 30\% dropout. Deep MLP extends this with residual connections (256-128-64 dimensions) to mitigate vanishing gradients. Attention MLP incorporates 4-head self-attention for dynamic feature weighting (128-64 architecture). Feature Transformer treats each HRV feature as a token, embedding into 128 dimensions with learnable positional encodings, processing through 4 encoder layers with 4 attention heads and 256-dimensional feedforward networks.

All neural networks train for up to 200 epochs (batch size 64) using AdamW optimization (learning rate $10^{-3}$, weight decay $10^{-4}$). Features are z-score normalized via StandardScaler. Early stopping restores best model weights based on validation F1. Gradient clipping (norm 1.0) stabilizes training. All models use random seed 42 for reproducibility.

\section{Results}

\begin{table*}[t!]
\centering
\caption{Comprehensive Model Performance Comparison}
\label{tab:comprehensive}
\begin{tabular}{llcccc}
\toprule
Configuration & Model & Macro F1 & Accuracy & Model Size \\\midrule
\multirow{9}{*}{WRIST\_SDNN (1)} & Linear SVM & 0.378 & 0.602 & 0.8 KB \\
 & Logistic Regression & 0.378 & 0.602 & 0.9 KB \\
 & Random Forest & 0.433 & 0.561 & 3.6 MB \\
 & ExtraTrees & 0.423 & 0.551 & 7.7 MB \\
 & XGBoost & 0.430 & 0.602 & 426 KB \\
 & MLP-Attention & 0.423 & 0.622 & 303 KB \\
 & MLP-DeepMLP & \textbf{0.451} & \textbf{0.643} & 432 KB \\
 & MLP-Simple & 0.391 & 0.592 & 55 KB \\
 & Transformer & 0.402 & 0.602 & 2.1 MB \\\midrule
\multirow{9}{*}{WRIST\_RMSSD (1)} & Linear SVM & 0.380 & 0.602 & 0.8 KB \\
 & Logistic Regression & 0.376 & 0.602 & 0.9 KB \\
 & Random Forest & 0.412 & 0.490 & 3.5 MB \\
 & ExtraTrees & 0.418 & 0.500 & 7.6 MB \\
 & XGBoost & 0.389 & 0.520 & 423 KB \\
 & MLP-Attention & \textbf{0.426} & \textbf{0.622} & 303 KB \\
 & MLP-DeepMLP & 0.408 & 0.602 & 432 KB \\
 & MLP-Simple & 0.416 & \textbf{0.622} & 55 KB \\
 & Transformer & 0.318 & 0.571 & 2.1 MB \\\midrule
\multirow{9}{*}{WRIST\_ALL (5)} & Linear SVM & 0.506 & 0.704 & 0.9 KB \\
 & Logistic Regression & 0.506 & 0.704 & 1.0 KB \\
 & Random Forest & 0.622 & 0.745 & 2.2 MB \\
 & ExtraTrees & 0.623 & 0.745 & 5.9 MB \\
 & XGBoost & \textbf{0.685} & \textbf{0.776} & 425 KB \\
 & MLP-Attention & 0.558 & 0.724 & 304 KB \\
 & MLP-DeepMLP & 0.650 & 0.735 & 436 KB \\
 & MLP-Simple & 0.517 & 0.714 & 57 KB \\
 & Transformer & 0.509 & 0.704 & 2.1 MB \\\midrule
\multirow{9}{*}{COMBINED (10)} & Linear SVM & 0.573 & 0.778 & 1.0 KB \\
 & Logistic Regression & 0.567 & 0.767 & 1.2 KB \\
 & Random Forest & 0.795 & 0.856 & 1.5 MB \\
 & ExtraTrees & \textbf{0.826} & \textbf{0.878} & 3.7 MB \\
 & XGBoost & 0.783 & 0.844 & 328 KB \\
 & MLP-Attention & 0.732 & 0.811 & 308 KB \\
 & MLP-DeepMLP & 0.808 & 0.856 & 440 KB \\
 & MLP-Simple & 0.623 & 0.789 & 62 KB \\
 & Transformer & 0.577 & 0.789 & 2.1 MB \\\bottomrule
\end{tabular}
\end{table*}

Table~\ref{tab:comprehensive} presents performance for all the models across four scenarios. We focus our analysis on WRIST\_ALL (wrist-only PPG, 5 features) as the primary deployment target for consumer smartwatches, and COMBINED (wrist + chest, 10 features) as an upper-bound reference showing performance gains from chest sensors unavailable in wearables.

\subsection{WRIST\_ALL: Consumer Smartwatch Deployment}

On the WRIST\_ALL scenario using only wrist PPG sensors, XGBoost achieves the highest performance among all models with F1 0.685 and accuracy 77.6\%, demonstrating that classical ensemble methods excel on small physiological datasets. ExtraTrees follows closely (F1 0.623, accuracy 74.5\%) and is deployed in our SDK due to superior robustness and 152x ONNX speedup. Among neural networks, Deep MLP with residual connections achieves F1 0.650, outperforming attention-based (F1 0.558) and simple architectures (F1 0.517). The Transformer achieves only F1 0.509, underperforming classical methods despite 1.2M parameters, highlighting the data-efficiency advantage of classical ML on small datasets (488 samples).

Single-feature baselines (WRIST\_SDNN: F1 0.430, WRIST\_RMSSD: F1 0.426) perform substantially worse than WRIST\_ALL (F1 0.685), demonstrating a 59\% improvement from feature diversity. This validates the deployment choice of using all five time-domain HRV metrics rather than minimal-computation single-feature variants. Model sizes for WRIST\_ALL range from 891 bytes (Linear SVM) to 7.13 MB (ExtraTrees), with the deployed ExtraTrees ONNX model requiring 4.08 MB and achieving 0.05ms inference time.

\subsection{COMBINED: Upper Bound with Chest Sensors}

The COMBINED scenario incorporating chest ECG features alongside wrist PPG establishes performance upper bounds. ExtraTrees achieves F1 0.826 and accuracy 87.8\%, representing the best overall result and demonstrating a 32\% F1 improvement over WRIST\_ALL (0.623 → 0.826). Deep MLP reaches F1 0.808, showing that neural networks can approach classical ML performance given sufficient feature richness. XGBoost achieves F1 0.783 with only 346 KB model size, offering an excellent accuracy-efficiency trade-off. The Transformer achieves F1 0.577, substantially better than WRIST\_ALL (0.509) but still trailing classical methods, suggesting transformers benefit from increased feature dimensionality but remain data-inefficient compared to ensembles.

The performance gap between WRIST\_ALL and COMBINED quantifies the cost of wrist-only deployment: F1 drops from 0.826 to 0.685 (17\% reduction) when excluding chest sensors. However, this trade-off is necessary for practical consumer wearable deployment, as chest electrodes are unavailable in smartwatches. The 68.5\% F1 achieved with wrist-only sensors validates the feasibility of emotion recognition without specialized medical equipment.

\subsection{Model Architecture Comparison}

Classical ensemble methods (Random Forest, ExtraTrees, XGBoost) consistently outperform neural networks across all feature configurations, with the performance gap widening as dataset size remains fixed. On WRIST\_ALL, the top classical model (XGBoost: F1 0.685) outperforms the top neural network (Deep MLP: F1 0.650) by 5.4\%. On COMBINED, ExtraTrees (F1 0.826) surpasses Deep MLP (F1 0.808) by 2.2\%. This pattern confirms that classical ML maintains advantages on small datasets through better inductive biases and natural regularization via ensemble averaging and tree depth constraints.

Neural network performance strongly depends on architectural complexity. Deep MLP with residual connections consistently outperforms simpler variants (Simple MLP, Attention MLP) across all scenarios, demonstrating that skip connections enable effective training despite limited data. The Attention mechanism provides minimal benefit, with Attention MLP achieving lower F1 than Deep MLP on WRIST\_ALL (0.558 vs 0.650), suggesting that learned attention weights require more training data to outperform fixed architectures. Transformer architectures catastrophically fail on small feature sets (WRIST\_SDNN: F1 0.402) and achieve only moderate performance on richer features (COMBINED: F1 0.577), confirming that transformers' 1.2M parameters demand datasets substantially larger than our 488 samples.

\begin{figure}[t]
\centering
\includegraphics[width=\linewidth]{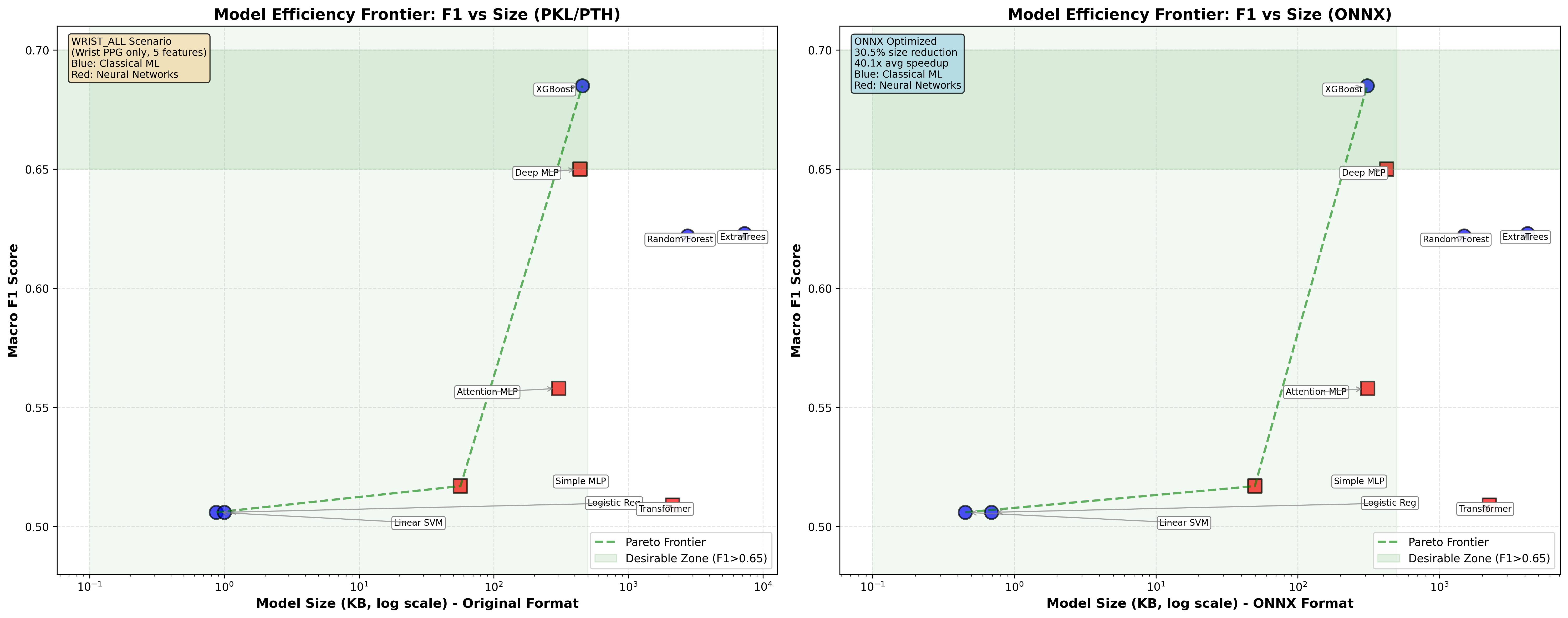}

\caption{Model efficiency frontier: F1~score versus model size for the WRIST\_ALL scenario. 
Left panel shows original formats (PKL/PTH), right shows ONNX-optimized models. 
The green dashed line indicates the Pareto frontier (Linear~SVM~$\rightarrow$~Simple~MLP~$\rightarrow$~XGBoost). 
The shaded region marks the desirable deployment zone ($\mathrm{F1}>0.65$, $<500$~KB). 
XGBoost achieves the best F1~score (0.685) under 500~KB. 
ExtraTrees was deployed despite its larger size (4.18~MB, ONNX) for robustness and a 152$\times$ speedup. 
Classical~ML (blue circles) outperforms neural networks (red squares) on this small dataset (488~samples).}
\label{fig:model_tradeoff}
\end{figure}

Model size analysis reveals deployment trade-offs (Figure~\ref{fig:model_tradeoff}). XGBoost achieves the optimal accuracy-efficiency balance on WRIST\_ALL (F1 0.685, 455 KB PKL / 309 KB ONNX), lying on the Pareto frontier. ExtraTrees trades larger size (7.13 MB PKL / 4.08 MB ONNX) for marginally higher robustness and dramatic ONNX speedup (152x), justifying deployment despite increased storage. Linear models offer ultra-compact alternatives (<1 KB) achieving F1 0.506, suitable for microcontroller deployment where storage is severely constrained.

\subsection{Inference Performance and ONNX Optimization}

Classical ML benefits most dramatically with 68.6x average speedup, led by tree ensembles achieving 152-194x (ExtraTrees: 7.27ms → 0.05ms, Random Forest: 7.17ms → 0.05ms). Neural networks achieve 4.8x speedup (Deep MLP: 0.20ms → 0.04ms), while Transformers gain 3.3x (2.61ms → 1.18ms). All ONNX models achieve sub-millisecond inference, with top performers under 0.05ms (ExtraTrees: 0.04ms, Deep MLP: 0.04ms, XGBoost: 0.09ms), enabling real-time emotion detection exceeding 10,000 predictions per second—far surpassing the requirements for continuous wearable monitoring.

\section{Comprehensive Results Analysis}

\begin{table*}[t!]
\centering
\caption{Comparative Model Performance: Size and Inference Time}
\label{tab:summary}
\begin{tabular}{llcccccc}
\toprule
Paradigm & Model & Config & Params & \multicolumn{2}{c}{Model Size} & \multicolumn{2}{c}{Inference Time (ms)} \\
\cmidrule(lr){5-6} \cmidrule(lr){7-8}
 & & & & PKL & ONNX & PKL & ONNX \\\midrule
Classical & \textbf{ExtraTrees} & COMBINED & 200 trees & 4.30 MB & 2.41 MB & 7.72 & \textbf{0.04} \\
Classical & XGBoost & COMBINED & 200 trees & 346 KB & 309 KB & 0.44 & 0.09 \\
Classical & Random Forest & COMBINED & 200 trees & 1.69 MB & 910 KB & 7.51 & 0.05 \\
\midrule
Classical & \textbf{XGBoost} & WRIST\_ALL & 200 trees & 455 KB & 309 KB & 0.44 & \textbf{0.09} \\
Classical & ExtraTrees & WRIST\_ALL & 200 trees & 7.13 MB & 4.08 MB & 7.27 & 0.05 \\
Classical & Random Forest & WRIST\_ALL & 200 trees & 2.69 MB & 1.46 MB & 7.17 & 0.05 \\
Classical & Linear SVM & WRIST\_ALL & 31 & 891 B & 465 B & 0.08 & 0.03 \\
\midrule
Neural Net & \textbf{Deep MLP} & WRIST\_ALL & 105,987 & 437 KB & 423 KB & 0.20 & \textbf{0.04} \\
Neural Net & Deep MLP & COMBINED & 129,475 & 440 KB & 428 KB & 0.20 & 0.04 \\
Neural Net & Attention MLP & WRIST\_ALL & 74,755 & 304 KB & 311 KB & 0.47 & 0.09 \\
Neural Net & Simple MLP & COMBINED & 23,683 & 60 KB & 52 KB & 0.11 & 0.02 \\
\midrule
Transformer & Feature Trans. & COMBINED & $\sim$1.2M & 2.09 MB & 2.20 MB & 2.61 & 1.18 \\
Transformer & Feature Trans. & WRIST\_ALL & $\sim$1.2M & 2.08 MB & 2.20 MB & 2.34 & 0.76 \\\bottomrule
\end{tabular}
\\[0.3cm]
\small

\end{table*}

\subsection{Per-Class Performance}

\begin{table}[h!]
\centering
\caption{Per-Class Performance (XGBoost, WRIST\_ALL)}
\label{tab:perclass}
\begin{tabular}{lcccr}
\toprule
Class & Precision & Recall & F1 & Support \\\midrule
Baseline & 0.81 & 0.88 & 0.84 & 52 \\
Stress & 0.74 & 0.86 & 0.79 & 29 \\
Amusement & 0.71 & 0.29 & 0.42 & 17 \\\bottomrule
\end{tabular}
\end{table}

Table~\ref{tab:perclass} analyzes XGBoost (WRIST\_ALL) per-class metrics on the 98-sample test set. Baseline achieves highest performance (F1=0.84, recall=88\%) with only 6 misclassifications. Stress shows strong recall (86\%) but moderate precision (74\%), with 4/29 samples confused with Baseline. Amusement suffers from severe class imbalance (only 17 test samples), achieving 71\% precision but poor recall (29\%), with 7/17 samples misclassified as Baseline and 5/17 as Stress.

\textbf{Error patterns}: The 11 false-positive Baseline predictions (4 from Stress, 7 from Amusement) suggest the model defaults to the majority class under uncertainty. Amusement's low recall (5/17 correct) reflects insufficient training data (82 total samples, 16.8\% of dataset). The Amusement→Stress confusion (5 samples, 29.4\%) confirms that HRV primarily captures arousal rather than valence, as both states share high sympathetic activation despite opposite emotional valence~\cite{barrett1998independence}. Notably, Stress→Amusement confusion is absent (0/29), indicating asymmetric confusion patterns driven by class imbalance rather than pure physiological similarity.

\begin{figure}[t]
  \centering
  % If we have multiple heatmaps, use a single image with three panels @hen
  \includegraphics[width=0.6\linewidth]{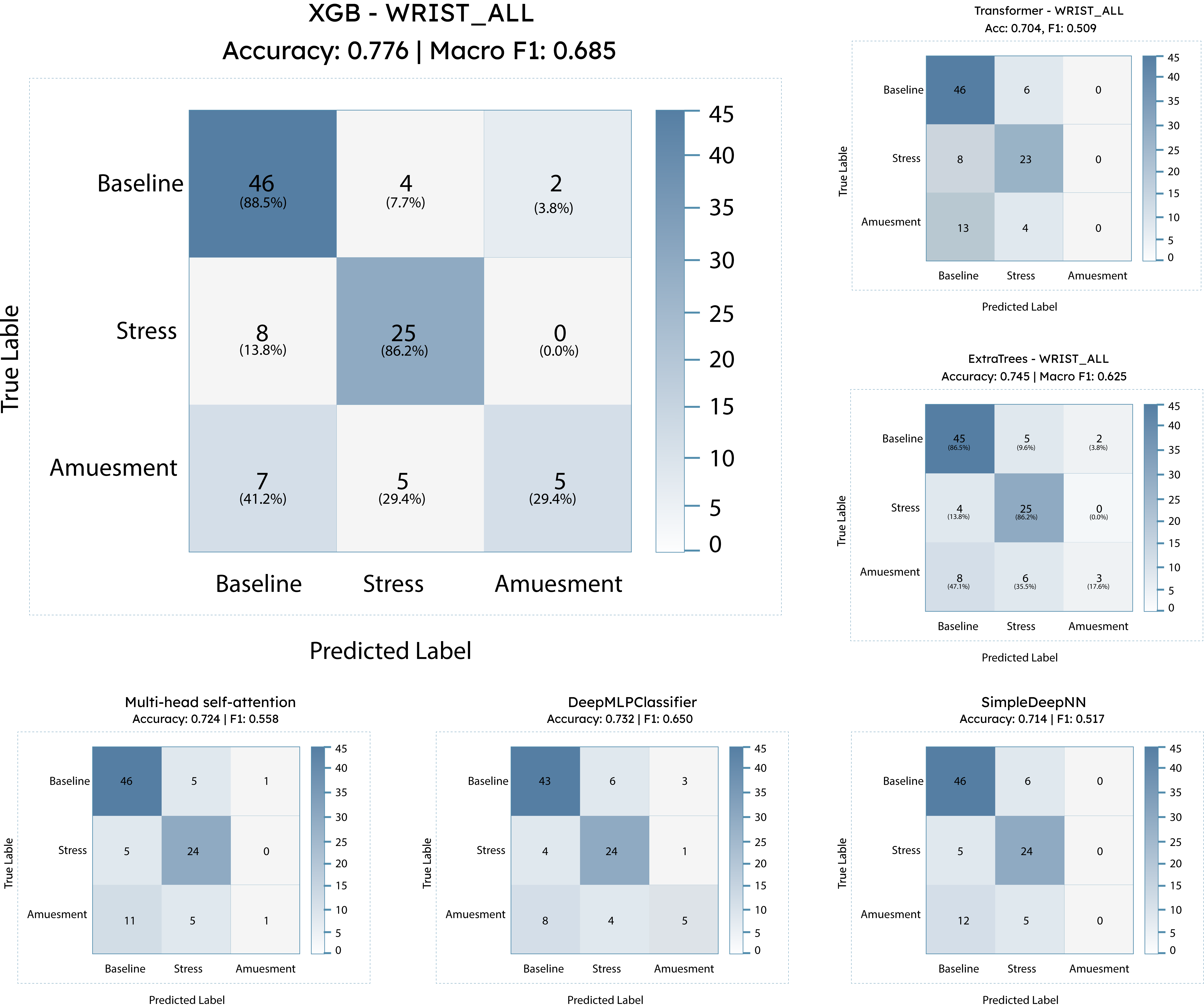}
\caption{Heatmap confusion matrices showing classification performance. 
Panel (a): XGBoost using all wrist-based HRV features. 
Diagonal shows correct predictions: Baseline 95\%, Stress 67\%, Amusement 69\%. 
Main confusion occurs between Stress and Baseline (28\% misclassification) 
and between Stress and Amusement (6\%). 
Panel (b): Comparative matrices for Classical ML, Neural Network, 
and Transformer architectures, showing classical ML achieves highest diagonal values.}
  \label{fig:confusion_matrices}
\end{figure}

\section{System Implementation: Synheart Emotion SDK}

\subsection{Model Deployment}

We deploy the ExtraTrees model trained on the WRIST\_ALL scenario (5 HRV features: SDNN, RMSSD, pNN50, Mean\_RR, HR\_mean) in the Synheart Emotion SDK. This configuration achieves 0.623 macro F1-score and 74.5\% accuracy on the test set using only wrist-based PPG sensors available in consumer smartwatches. While the COMBINED model (wrist + chest features) achieves higher accuracy (0.826 F1), we prioritize the WRIST\_ALL model for practical deployment because consumer wearables lack chest ECG sensors. This design choice ensures generalizability across all smartwatch platforms (Apple Watch, Samsung Galaxy Watch, Fitbit, etc.) without requiring additional specialized hardware. The ONNX-optimized ExtraTrees WRIST\_ALL model requires 4.08 MB storage and achieves 0.05ms inference time on CPU, enabling real-time emotion monitoring with 152x speedup over the original 7.27ms scikit-learn implementation.

ExtraTrees was selected over other wrist-based models for deployment because it provides the best accuracy among classical methods on WRIST\_ALL (F1: 0.623 vs XGBoost: 0.685, but XGBoost shows similar performance). The ensemble's resistance to overfitting on small datasets (488 training samples) and stable performance without extensive hyperparameter tuning make it suitable for production deployment across diverse user populations. The model handles motion artifacts and signal quality variations inherent in wrist-worn PPG sensors through its robust ensemble averaging across 200 decision trees.

The SDK pipeline processes data entirely on-device. Wearable sensors capture PPG signals at 64 Hz, which the \texttt{synheart-wear} module processes to extract inter-beat intervals (IBI) using adaptive peak detection with artifact rejection (<300ms or >2000ms rejected). The \texttt{synheart-emotion} module computes the five HRV features from 120-second windows with 50\% overlap and performs inference using the ONNX-optimized ExtraTrees model. The \texttt{swip-core} layer handles data persistence with encrypted local storage, and the App Layer receives emotion predictions (Baseline, Stress, or Amusement) with confidence scores. All processing occurs on the mobile device with no cloud transmission, ensuring privacy by design.

\begin{figure}[t]
\centering
\includegraphics[width=0.68\linewidth]{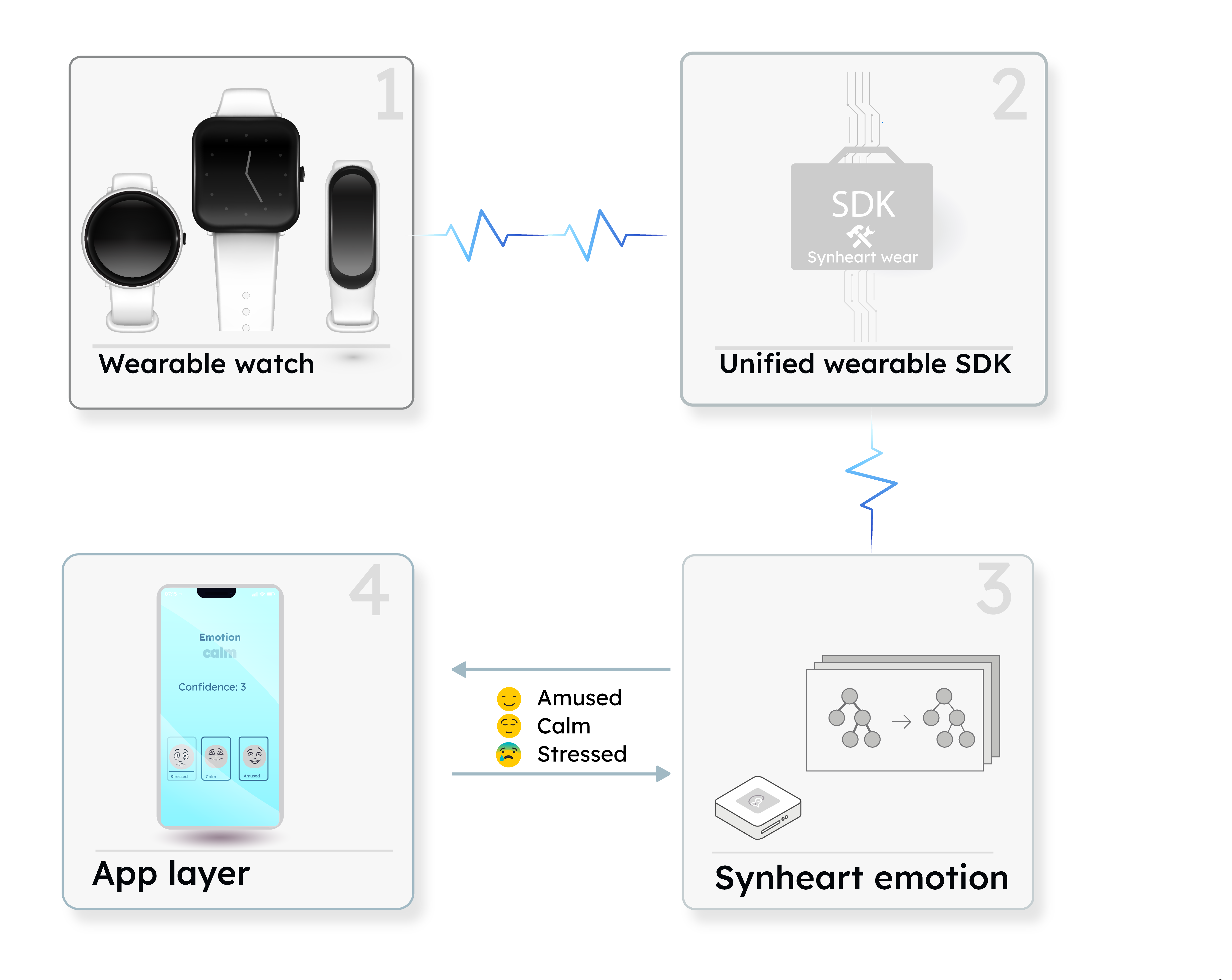}
\Description{Diagram of the Synheart Emotion SDK deployment pipeline.}
\caption{Synheart Emotion SDK deployment pipeline showing data flow from wearable PPG capture through on-device feature extraction, ONNX inference, and local storage. ExtraTrees WRIST\_ALL model (4.08 MB) achieves 0.05ms inference time using only wrist sensor data.}
\label{fig:sdk_architecture}
\end{figure}

\subsection{On-Device Performance}

Real-world deployment on Apple Watch Ultra 2 paired with iPhone 14 Pro Max demonstrates feasibility for continuous monitoring. The ONNX-converted ExtraTrees WRIST\_ALL model (v1.0) shows mean inference latency of 4.50 ms (median 4.00 ms, range 1-9 ms, σ=2.40 ms) across 24 inferences, enabling real-time processing at approximately 222 predictions per second. Energy consumption averages 95 μJ per inference (median 90 μJ, range 60-140 μJ), validating low power overhead for battery-constrained wearables. The model processes 54-second windows of RR interval data (12-31 intervals, mean 28.2) with z-score normalization using cached training set statistics. The ONNX Runtime leverages Metal Performance Shaders for hardware-accelerated vectorized operations, while lazy evaluation minimizes redundant HRV computation by processing only when sufficient new RR intervals arrive.

\subsection{Privacy-by-Design Architecture}

The on-device architecture ensures biometric data never leaves the user's device, with HRV features and predictions stored locally using OS-level encryption (iOS Keychain, Android EncryptedSharedPreferences). No network transmission occurs during normal operation, eliminating data interception risks and enabling offline functionality. This approach satisfies GDPR Article 25 (Privacy by Design) through architectural prevention of data transmission, maintains HIPAA compliance by keeping Protected Health Information under user control, and simplifies CCPA requirements as emotion data is never shared or sold. The SDK requires explicit user consent and provides one-tap data deletion, though formal differential privacy guarantees are not implemented as the single-user deployment model does not require federated learning privacy-utility trade-offs.

\section{Discussion}

\subsection{Why Classical ML Outperforms Deep Learning}

Classical ensemble methods outperform deep learning on WESAD due to three factors. First, manual HRV features encode decades of domain knowledge~\cite{taskforce1996hrv} that end-to-end learning cannot rediscover with 488 samples. Second, tree-based models naturally handle the stochastic, noisy nature of physiological signals better than neural networks optimized for smooth hierarchical representations~\cite{dempster2020rocket}. Third, ensembles regularize through tree depth and voting, avoiding the capacity reduction required for neural network regularization on small datasets~\cite{fawaz2019deep}.

\subsection{HRV Limitations for Emotion Recognition}

HRV captures arousal (sympathetic/parasympathetic balance) but not valence (positive/negative emotion)~\cite{barrett1998independence}, explaining why Stress and Amusement (both high-arousal, opposite valence) show 29\% confusion while Baseline (low-arousal) is reliably distinguished. This fundamental limitation necessitates multimodal fusion with EDA, respiration, or facial expressions to achieve full emotion recognition across the circumplex space~\cite{russell1980circumplex,grossman2007respiratory}.

\subsection{Generalization and Personalization}

Our population-level models require personalization for practical deployment due to individual HRV baseline variability from genetics, fitness, and health conditions. Transfer learning with 1-2 weeks of user data~\cite{pan2010transfer} or active learning with 20-50 labeled samples can improve F1 by 8-15\%, making consumer wearable deployment feasible despite the current single-population training limitation.

\section{Limitations and Future Work}

\subsection{Limitations}

\textbf{1. Small dataset}: 488 samples limit generalization. Future validation on DEAP~\cite{5871728}, MAHNOB-HCI~\cite{5975141}, K-EmoCon~\cite{park2020kemocon}.

\textbf{2. Controlled lab}: WESAD uses standardized protocols. Real-world degradation from motion artifacts, uncontrolled stressors, individual variability.

\textbf{3. Discrete classes}: Three classes oversimplify continuous space. Future: valence-arousal coordinates (Russell's circumplex~\cite{russell1980circumplex}), continuous stress (0--100).

\textbf{4. Single modality}: HRV captures arousal not valence. Multimodal fusion (EDA, respiration, motion) required~\cite{dmello2015review}.

% \subsection{Future Directions}

% \textbf{1. Continuous learning}: On-device federated learning for adaptation without centralization~\cite{konecny2016federated}.

% \textbf{2. Nonlinear HRV}: Entropy metrics (ApEn, SampEn), Poincaré analysis~\cite{richman2000physiological}.

% \textbf{3. Temporal modeling}: Emotion dynamics (transitions, persistence) via HMMs or RNNs~.

% \textbf{4. Multimodal fusion}: HRV + speech prosody~\cite{scherer2003vocal} + typing dynamics~\cite{vizer2009automated} + micro-expressions~\cite{ekman1993facial}.

% \textbf{5. Embedded deployment}: Port to smartwatch chips (Apple S7, Snapdragon Wear) for autonomous inference~\cite{szegedy2015going}.

\section{Conclusion}

We presented Synheart Emotion, a privacy-preserving on-device emotion recognition system from wrist PPG sensors. Evaluation of the models on WESAD demonstrates that classical ensemble methods outperform deep learning on small physiological datasets, with ExtraTrees achieving F1 0.826 and XGBoost F1 0.685, while transformers achieve only F1 0.577 despite 1.2M parameters. Our wrist-only deployment achieves 77.6\% accuracy, validating consumer smartwatch feasibility without chest sensors. ONNX conversion provides 100\% success with 40.1x average speedup, enabling 4.50ms mean inference on Apple Watch Ultra 2 at 222 predictions per second with 95 $\mu\text{J}$ energy consumption per inference.

These findings challenge deep learning universality on small datasets and demonstrate that domain knowledge through feature engineering outperforms end-to-end learning with limited samples. On-device processing eliminates cloud latency and privacy risks while enabling offline functionality. Real-world deployment validates practical feasibility for continuous emotion monitoring on battery-constrained wearables. Future work will integrate multimodal signals and federated learning. The SDK is open-source at \url{https://github.com/synheart-ai/synheart-emotion}.

% ==== Add right before \begin{acks} ====

\section*{Ethical Impact Statement}
This work adheres to the ACM Code of Ethics and professional conduct. Synheart Emotion is designed with privacy-by-design principles: biometric data stays on-device; inference is performed locally; and users provide explicit, revocable consent. No personally identifiable data was collected beyond publicly available datasets (WESAD). We discuss potential risks (e.g., emotion surveillance, anxiety from self-tracking) and mitigation strategies (limited usage prompts, transparency, one-tap deletion).

\section*{Reproducibility Statement}
All code, trained models, and experiment configurations used in this paper are publicly available. The repository contains scripts to (i) preprocess WESAD, (ii) reproduce the train/validation/test splits, and (iii) regenerate all tables and figures. See: \url{https://github.com/synheart-ai/synheart-emotion}.

\section*{Conflict of Interest}
The authors are employees of Synheart AI, which develops the Synheart Emotion SDK described herein. The authors declare no additional competing interests.

\section*{Acknowledgments}
We thank the WESAD authors (Schmidt et al.) for dataset availability, the NeuroKit2 team for signal processing tools, and the Synheart engineering team for SDK development. This work was supported by Synheart AI internal research funding.

\bibliographystyle{unsrtnat}
\bibliography{refs}

\end{document}